%% file: main.tex
\documentclass[10pt,twocolumn,letterpaper]{article}

\usepackage{cvpr}              
\input{preamble}
\definecolor{cvprblue}{rgb}{0.21,0.49,0.74}
\usepackage[pagebackref,breaklinks,colorlinks,allcolors=cvprblue]{hyperref}
\usepackage{multirow}
\usepackage{colortbl}
\usepackage{graphicx}
\definecolor{mygray}{gray}{0.5}
\usepackage{color}
\usepackage{tabularx}
\usepackage{makecell}

\def\paperID{} 
\def\confName{CVPR}
\def\confYear{2026}

\title{SWIFT: Sliding Window Reconstruction for Few-Shot Training-Free \\ Generated Video Attribution}

\author{
Chao Wang\textsuperscript{\rm 1}, 
Zijin Yang\textsuperscript{\rm 1}, 
Yaofei Wang\textsuperscript{\rm 2}, 
Yuang Qi\textsuperscript{\rm 1}, 
Weiming Zhang\textsuperscript{\rm 1}, 
Nenghai Yu\textsuperscript{\rm 1}, 
Kejiang Chen\textsuperscript{\rm 1,}\thanks{Corresponding author.} \\
\textsuperscript{\rm 1}University of Science and Technology of China \quad \textsuperscript{\rm 2}Hefei University of Technology \\
\href{chaowang0708@mail.ustc.edu.cn}{\tt chaowang0708@mail.ustc.edu.cn} \quad \href{chenkj@ustc.edu.cn}{\tt chenkj@ustc.edu.cn}
}

\begin{document}

\maketitle
\input{sec/0_Abstract} 
\input{sec/1_Intro}
\input{sec/2_Related_Work} 
\input{sec/4_Method}

\input{sec/5_Exper}
\input{sec/6_Limitation}
\input{sec/7_Conclusion}
\input{sec/8_Ack}
{
    \small
    \bibliographystyle{ieeenat_fullname}
    \bibliography{main}
}



\end{document}

%% file: sec/0_Abstract.tex
\begin{abstract}

\vspace{-0.2cm} 

Recent advancements in video generation technologies have been significant, resulting in their widespread application across multiple domains. However, concerns have been mounting over the potential misuse of generated content. Tracing the origin of generated videos has become crucial to mitigate potential misuse and identify responsible parties. Existing video attribution methods require additional operations or the training of source attribution models, which may degrade video quality or necessitate large amounts of training samples. To address these challenges, we define for the first time the ``few-shot training-free generated video attribution" task and propose SWIFT, which is tightly integrated with the temporal characteristics of the video. By leveraging the ``Pixel Frames(many)$\leftrightarrow$Latent Frame(one)" temporal mapping within each video chunk, SWIFT applies a fixed-length sliding window to perform two distinct reconstructions: normal and corrupted. The variation in the losses between two reconstructions is then used as an attribution signal. We conducted an extensive evaluation of five state-of-the-art (SOTA) video generation models. Experimental results show that SWIFT achieves over 90\% average attribution accuracy with merely 20 video samples across all models and even enables zero-shot attribution for HunyuanVideo, EasyAnimate, and Wan2.2. Our source code is available at this repository\footnote{\url{https://github.com/wangchao0708/SWIFT}}.

\vspace{-0.4cm}



\end{abstract}

%% file: sec/1_Intro.tex
\section{Introduction}
\label{Introduction}

Video generation technology has recently achieved significant advancements, with Latent Diffusion Models (LDMs)~\cite{Cog, MS, StepVideo, VC2, Animatediff, OpenSora, LTX} emerging as the dominant paradigm in this field due to their exceptional capability in producing highly realistic content. The advent of Sora~\cite{Sora} in 2024 marked a significant breakthrough in video generation, accelerating the widespread adoption of this technology. Subsequently, open-source models like HunyuanVideo~\cite{HYV} and Wan2.1~\cite{Wan2.1} were released. 

However, the rapid development of video generation technology has simultaneously intensified concerns regarding its potential for misuse~\cite{misuse-1, VGM, misuse-3}. Malicious actors could use generated videos to infringe on intellectual property (IP)~\cite{IP} or spread disinformation~\cite{earthquake}, threatening the security of the information ecosystem. As a result, reliable source attribution (identifying which generator produced a given synthetic video) has become essential for determining the responsible parties behind generated content.

\begin{figure}[]
\centering
\includegraphics[width=\linewidth]{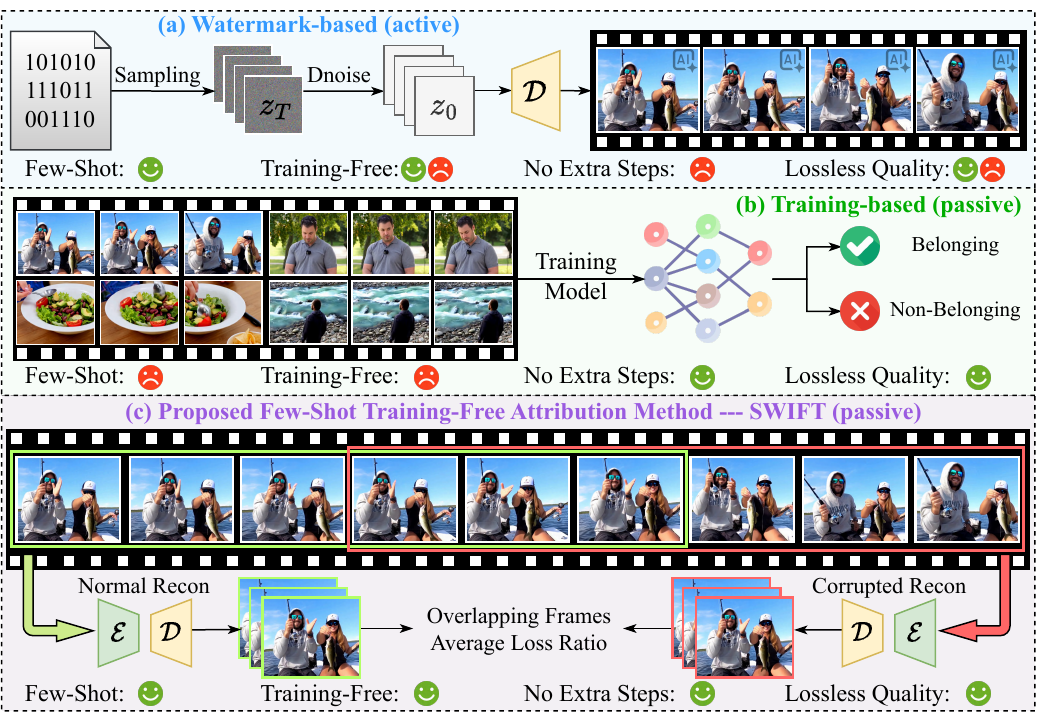}
\vspace{-0.7cm}
\caption{Comparison of existing attribution methods and SWIFT: (a) Watermark-based active attribution; (b) Training-based passive attribution; (c) SWIFT: few-shot training-free passive attribution.}
\label{Method-Comparison}
\vspace{-0.55cm} 
\end{figure}

Existing methods for generated video attribution can be divided into two types: active attribution and passive attribution. The most representative approach within active attribution is watermark-based methods, which embed model ownership information during~\cite{Videoshield, Videomark} or after~\cite{water-post-1, water-post-2} the generation process for traceability, as shown in \cref{Method-Comparison}(a). Passive attribution typically employs training-based methods~\cite{VGM}, where a source-tracing model is trained to perform attribution, as shown in \cref{Method-Comparison}(b). Although both methods effectively achieve the attribution goal, they may result in a decline in video quality or require a large number of training samples, making them impractical in many contexts.


In contrast, few-shot training-free passive generated video attribution methods show greater practical potential. While no mature solutions are currently available in this area, breakthroughs~\cite{RONAN, LatentTracer, AEDR} in image attribution tasks provide valuable insights. These methods typically use reconstruction error as an attribution signal. For instance, RONAN~\cite{RONAN} and LatentTracer~\cite{LatentTracer} initialize from a random starting point or a VAE-encoded latent representation, respectively, and perform reconstruction through gradient-based optimization. However, gradient-based optimization imposes significant computational overhead and lengthy execution times. By comparison, AEDR~\cite{AEDR} reconstructs inputs solely through the target model’s VAE and determines attribution based on the latent feature consistency of belonging images. This design not only significantly reduces computational overhead but also improves attribution accuracy.


Despite their success in image modality, reconstruction-based generated image attribution methods~\cite{RONAN, LatentTracer, AEDR} exhibit a marked decline in attribution accuracy when transferred to the video modality (as detailed in the supplementary material). This degradation stems from the complex temporal characteristics inherent in video data. These methods primarily emphasize spatial consistency in reconstruction while overlooking the crucial temporal coherence constraints required to effectively handle sequence-dependent perturbations. This brings us to a crucial question: How can we develop an effective reconstruction strategy that deeply captures the temporal characteristics of generated videos?

\begin{figure}[]
\centering
\includegraphics[width=\linewidth]{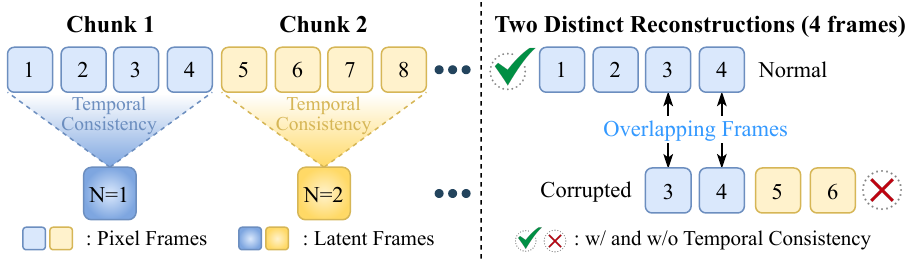}
\vspace{-0.7cm}
\caption{The 3D VAE performs up and down sampling operations along the temporal dimension (Temporal Compression Ratio = 4) and executes two distinct reconstructions (Normal and Corrupted).}
\label{3D VAE}
\vspace{-0.5cm}  
\end{figure}

We observe that SOTA video generation models~\cite{Sora, HYV, EA, Wan2.1, LTX, StepVideo, OpenSora, Cog, MS} often employ 3D VAEs to handle the high computational demands of generation. Unlike image or earlier video VAEs, 3D VAEs incorporate temporal up and down sampling within the latent space. This mechanism naturally forms temporal consistency that aligns with the VAE distribution within each chunk, a mapping relationship of ``\textbf{Pixel Frames(many)$\leftrightarrow$Latent Frame(one)}" (see \cref{3D VAE}). When pixel frames within a chunk undergo displacement, the temporal consistency becomes corrupted. Compared with normal reconstruction, corrupted reconstruction that breaks temporal consistency leads to a significant increase in loss (see \cref{3D VAE,Framework}).

Motivated by this insight, we propose a \textbf{S}liding \textbf{Wi}ndow reconstruction-based \textbf{F}ew-shot \textbf{T}raining-free video attribution method, named \textbf{SWIFT}. Belonging videos are decoded by the target model's VAE, so each chunk carries a temporal mapping consistent with the VAE distribution, which is absent in non-belonging videos. The core idea is to create a significant difference between two reconstructions by corrupting the temporal consistency, which can serve as an attribution signal. Specifically, SWIFT uses sliding window reconstruction, where the first reconstruction starts at the beginning of the video with a fixed-length window, and the second shifts the window backward to disrupt temporal mapping (see \cref{Framework}). For belonging videos, the first reconstruction conforms to the VAE distribution, while the second is deliberately corrupted, leading to a significant increase in loss. In contrast, non-belonging videos, which do not satisfy the VAE distribution, exhibit a loss ratio for the overlapping frames (see \cref{3D VAE}) close to 1 between two reconstructions. Furthermore, we introduce kernel density estimation~\cite{KDE} for adaptive threshold selection, which does not rely on prior knowledge of the underlying data distribution, thereby enhancing generalization capabilities. 

Our contributions are summarized as follows:
\begin{itemize}
    \item \textbf{New Paradigm:} Our work systematically uncovers the fundamental bottlenecks in current generated video attribution and establishes a novel research paradigm by first formally defining the ``few-shot training-free generated video attribution" task for reliable source tracing.
    \item \textbf{Framework Design:} To overcome these bottlenecks, we propose SWIFT, the first few-shot training-free video attribution framework that explicitly leverages the temporal mapping relationship inherent in modern 3D-VAEs to achieve efficient and reliable origin attribution.
    \item \textbf{Experimental Validation:} Extensive experiments on five SOTA video generation models demonstrate that SWIFT can effectively perform source attribution, achieving an average accuracy of 94\%, and reaching approximately 90\% accuracy in zero-shot scenarios for certain models.
\end{itemize}

%% file: sec/2_Related_Work.tex
\section{Related Work}

\noindent
\textbf{Video Generation Models.} In recent years, video generation models~\cite{Cog, EA, HYV, LTX, StepVideo, OpenSora} have advanced significantly, particularly in Text-to-Video (T2V) and Image-to-Video (I2V) tasks using latent diffusion models. Early models~\cite{Animatediff, MS, SVD, VC2} employed 2D VAEs to transform between pixel and latent spaces and used U-Net~\cite{U-net} to perform the denoising process, but often produced videos with issues like flickering and object deformation, limiting their widespread adoption. The breakthrough in bringing video generation technology into the mainstream came with the release of Sora~\cite{Sora}, which is capable of generating long and high-definition videos that align with physical principles. This milestone underscored the pivotal role of architectures that combine 3D VAE with Diffusion Transformer (DiT)~\cite{DiT} in video generation. As the resolution and frame count of generated videos improved, the computational resources required for training and inference surged dramatically. Consequently, most of the SOTA models incorporate 3D VAEs to compress spatial and temporal dimensions, optimize denoising processes, and accelerate generation speeds. Following the 3D VAEs and DiT paradigm, recent models such as EasyAnimate~\cite{EA}, Wan2.1~\cite{Wan2.1}, and HunyuanVideo~\cite{HYV} have achieved significant breakthroughs in both video quality and temporal consistency.

\noindent
\textbf{Detection of Generated Videos.} With the progress in video generation technology~\cite{EA, HYV, LTX, StepVideo, Cog, VC2}, the need for reliable detection has become increasingly pressing. Current detection methods have made notable strides, evolving from early models focused on detecting facial artifacts~\cite{DeepFake-1, DeepFake-2, DeepFake-3} to more general video detection approaches~\cite{UNITE, DeMamba, MM-Det, VGM, BusterX++}. The early general detection method, DeMamba~\cite{DeMamba}, introduced the unique Mamba module to analyze temporal and spatial inconsistencies for video detection. Subsequently, UNITE~\cite{UNITE} further improved detection by leveraging domain-agnostic features through SigLIP-So400M~\cite{SigLIP} and combining Adversarial and Cross-Entropy losses. Although these methods demonstrate promising performance in generated video detection, they do not address the more challenging task of video origin attribution, which is the focus of this paper.


\noindent
\textbf{Origin Attribution of Generated Videos.} Research on the attribution of generated videos has lagged behind detection technologies. Existing methods can be divided into two main categories: active attribution and passive attribution. Active attribution typically relies on watermark-based methods, which involve embedding model ownership information during~\cite{Videoshield, Videomark} or after~\cite{water-post-1, water-post-2, water-post-3, water-post-4} the generation process to facilitate copyright identification and source tracing. However, these methods require additional embedding operations or complex time-varying keys, which may negatively affect the visual quality or introduce substantial deployment challenges, making it unacceptable to both model owners and users. In contrast, passive attribution typically employs training-based methods~\cite{VGM}, where a generation source tracking model is trained to perform the attribution. This approach not only requires vast amounts of training data and high costs but also necessitates re-training the model whenever a new generation source emerges. Compared to these methods, SWIFT overcomes these limitations by employing a few-shot training-free strategy.


%% file: sec/4_Method.tex
\section{Method}
\label{Method}

\begin{figure*}[h]
    \centering
    \includegraphics[width=0.99\linewidth]{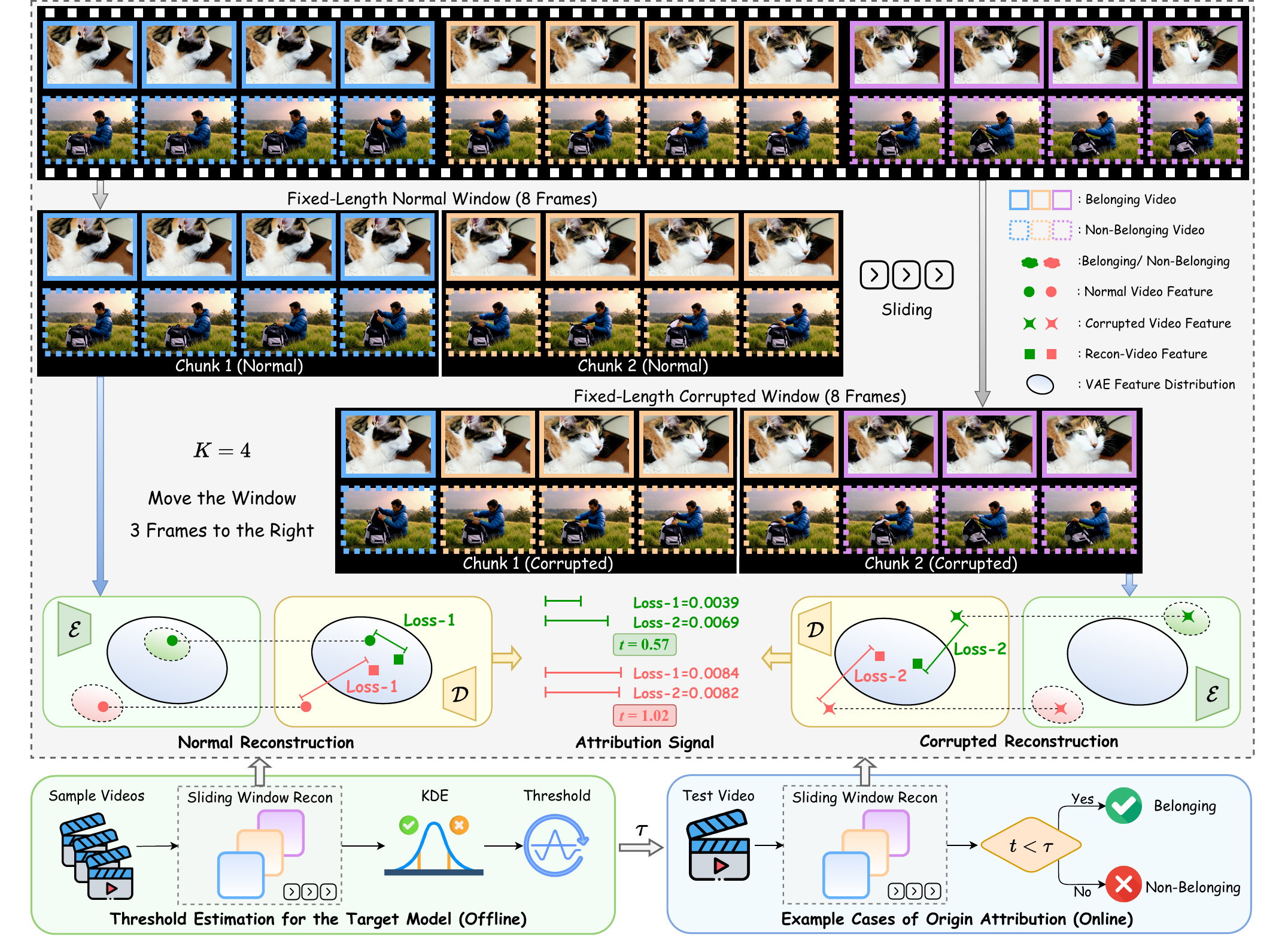}
    \vspace{-0.2cm}
    \caption{The framework of SWIFT consists of three key modules: determination of a fixed-length sliding window, normal and corrupted reconstruction, and threshold determination. SWIFT uses the average loss ratio of overlapping frames between two differential reconstructions as the attribution signal. This signal is then compared with the threshold obtained through KDE to derive the final attribution result.}
    \label{Framework}
    \vspace{-0.4cm} 
\end{figure*}

This section provides a detailed description of SWIFT. The process begins by determining the size of the sliding window, followed by two rounds of reconstruction, during which the loss for overlapping frames in both normal and corrupted reconstructions is computed. Then, the average loss ratio across these frames is calculated as the attribution signal. Finally, the threshold is determined using KDE, and the signal is compared against it to produce the result. The overall framework of SWIFT is shown in \cref{Framework}.

\subsection{Concept Definition}
\label{Concept_Definition}

We aim to act as an auditor to achieve efficient and reliable generated video attribution through a few-shot training-free approach. For convenience in discussion, we first define the concepts of belonging and non-belonging videos, and then introduce the auditor’s attribution goal and capability.

\noindent
\textbf{Definition 3.1 (Belonging vs Non-Belonging Videos):} 
\label{def-1}
Given a video generation model $\mathcal{M}: \mathcal{I} \rightarrow \mathcal{X}_\mathcal{M}$, where $\mathcal{I}$ is the input space and $\mathcal{X}_\mathcal{M} \subset \mathcal{X}$ denotes the output space consisting of videos generated by $\mathcal{M}$, with $\mathcal{X}$ representing the full video space. For a test video $x \in \mathcal{X}$, we say that $x$ belongs to model $\mathcal{M}$ if and only if $x \in \mathcal{X}_\mathcal{M}$; otherwise, if $x \notin \mathcal{X}_\mathcal{M}$, we say that $x$ does not belong to model $\mathcal{M}$.

\noindent
\textbf{Definition 3.2 (Auditor's Attribution Goal):} 
\label{def-2}
Given a test video $x$ and a target model $\mathcal{M}$, the auditor’s goal is to determine whether $x$ was generated by $\mathcal{M}$. This goal can be formalized as a binary classification function $B: \{x, \mathcal{M}\} \to \{0, 1\}$, where the function takes $x$ and $\mathcal{M}$ as inputs, and returns the final classification result (0 for belonging videos, and 1 for non-belonging videos). The classifier $B$ should distinguish not only between videos generated by $\mathcal{M}$ and those generated by other models, but also differentiate between belonging videos and real videos.

\noindent
\textbf{Definition 3.3 (Auditor's Capability):} 
\label{def-3}
The auditor has white-box access only to the autoencoder $\mathcal{R}$ of the target model $\mathcal{M}$. This assumption is both realistic and practical, particularly in scenarios where the auditor is the model owner. Under this setting, the auditor can use $\mathcal{R}$ to reconstruct the video but cannot control the model's training or generation process, nor can the auditor perform any post-processing on the generated videos. Furthermore, the auditor must implement the above function $B$ in a few-shot training-free manner. This differs from existing methods, which typically require watermark embedding during the generation stage (with white-box access to the entire model)~\cite{Videoshield, Videomark} or after generation~\cite{water-post-1,water-post-2}, or rely on large volumes of samples to train source-tracing models~\cite{VGM}. In contrast, SWIFT requires no additional operations and does not depend on large samples, thereby exhibiting greater feasibility and practical value in restricted-access scenarios.

\subsection{Fixed-Length Sliding Window}

SWIFT employs the sliding window mechanism to perform two distinct reconstructions of the test video $x$. The key lies in determining the window size and its shifting strategy. Let the test video $x$ consist of multiple consecutive frames, formally represented as $x = \{ F_1, F_2, \dots, F_{KN} \}$, where $K$ denotes the temporal compression ratio of the target model’s 3D VAE (typically 4 or 8), and $N$ represents the total number of chunks in the test video $x$. Each chunk contains $K$ consecutive frames, defined as:
\vspace{-0.1cm}
\begin{equation}
\label{eq:chunk_def}
C_i = \{ F_{K(i-1)+1}, \dots, F_{Ki} \}, \quad i = 1, \dots, N.
\end{equation}
To maximize the visual information captured within a window, the window size is set to $K(N-1)$ frames, corresponding to $N-1$ chunks. The sliding window moves along the frame sequence with a stride of one frame, and the offset is denoted by $j$, where $ j = 0, 1, \dots, K $. The resulting series of windows can be expressed as:
\vspace{-0.1cm}
\begin{equation}
W_j = \{ F_{j+1}, F_{j+2}, \dots, F_{K(N-1)+j} \} .
\end{equation}
As $j$ increases by one, the entire window shifts forward by a single frame. Since video generation models~\cite{HYV, Wan2.1, EA} employ a 3D VAE that processes chunks of $K$ consecutive frames as the minimal temporal unit, the sliding windows under different offsets can be categorized into two types:
\vspace{-0.1cm}
\begin{align}
W_{\mathrm{Nor}}: j \bmod K = 0, \quad W_{\mathrm{Cor}}: j \bmod K \neq 0.
\end{align}
For normal windows, both the composition of video frames within each chunk and their positional alignment conform to \cref{eq:chunk_def}, thereby satisfying the temporal mapping of ``Pixel Frames(many)$\leftrightarrow$Latent Frame(one)". Therefore, $W_0$ and $W_K$ are classified as normal windows. In contrast, corrupted windows disrupt this correspondence, leading to the classification of $W_1$ through $W_{K-1}$ as corrupted windows.

To maximize the reconstruction discrepancy and obtain the most salient attribution signals, a precise approach involves identifying the most divergent combination by calculating the reconstruction loss for each window. Alternatively, a more intuitive and expedited method is to directly designate $W_0$ and $W_{K-1}$ as the normal and corrupted windows, respectively. This is because in $W_{K-1}$, every frame is shifted backward by the maximum offset of $K-1$ frames relative to $W_0$. This configuration alters both the frame composition within each chunk and the positional mapping of individual frames, thereby achieving a more pronounced corruption effect. For ease of discussion, we conduct the reconstruction analysis using $W_0$ and $W_{K-1}$.

\subsection{Normal and Corrupted Reconstruction}
\label{3-3}
After determining the normal and corrupted windows, we demonstrate the temporal mapping differences between belonging and non-belonging videos through the reconstructions. According to \cref{def-3}, as an auditor, we have white-box access to the 3D VAE $\mathcal{R}$ of the target model. Thus, we can perform reconstruction on both the normal window $W_0$ and the corrupted window $W_{K-1}$, expressed as follows:
\begin{equation}
W^*_{0} = \mathcal{R}(W_0) = \{ F_1^*, F_2^*, \dots, F_{K(N-1)}^* \},
\end{equation}
\begin{equation}
W^{**}_{K-1} = \mathcal{R}(W_{K-1}) = \{ F_K^{**}, F_{K+1}^{**}, \dots, F_{KN-1}^{**} \}.
\end{equation}
Here, $W^*_{0}$ and $W^{**}_{K-1}$ are the reconstructed normal and corrupted windows, respectively. We define the normal and corrupted reconstruction losses as follows:
\begin{equation}
\begin{aligned}
&\mathcal{L}_{\text{Nor}} = \mathcal{L}(W^*_{0}, W_0) \\
&= \{ \mathcal{L}(F_1^*, F_1), \mathcal{L}(F_2^*, F_2), \dots, \mathcal{L}(F_{K(N-1)}^*, F_{K(N-1)}) \} ,
\end{aligned}
\end{equation}
\begin{equation}
\begin{aligned}
&\mathcal{L}_{\text{Cor}} = \mathcal{L}(W^{**}_{K-1}, W_{K-1}) \\
&= \{ \mathcal{L}(F_K^{**}, F_K), \mathcal{L}(F_{K+1}^{**}, F_{K+1}), \dots, \mathcal{L}(F_{KN-1}^{**}, F_{KN-1}) \} ,
\end{aligned}
\end{equation}
where $\mathcal{L}$ is the loss metric, which we set as the Mean Squared Error (MSE); $\mathcal{L}_{\text{Nor}}$ and $\mathcal{L}_{\text{Cor}}$ represent the reconstruction losses for each frame in the normal and corrupted windows, respectively. SWIFT focuses on the differential changes in losses between these two rounds of reconstruction, and attribution signal $t$ is defined as the average loss ratio of the overlapping frames (see \cref{3D VAE,Framework}) from the two distinct reconstructions, which is expressed as:
\begin{equation}
t = \frac{1}{K(N-1) - K + 1} \sum_{i=K}^{K(N-1)} \frac{\mathcal{L}(F_i^*, F_i)}{\mathcal{L}(F_i^{**}, F_i)} .
\end{equation}

\subsection{Threshold Determination}
The attribution signal $t$ does not follow a consistent probability distribution across different models and may contain a few outliers (see supplementary material). Consequently, a distinct threshold $\tau$ must be determined for each model independently. To address this, we use kernel density estimation~\cite{KDE}, a non-parametric method that makes no assumptions about the underlying distribution and is inherently robust to outliers. Specifically, the threshold $\tau$ is derived from the cumulative distribution function estimated by KDE:
\begin{equation}
\tau = \inf \left\{ u \;\Bigm|\; \int_{-\infty}^{u} \frac{1}{S h} \sum_{q=1}^{S} G \left( \frac{y - t_q}{h} \right) \, dy \geq 1 - \alpha \right\} ,
\end{equation}
where $S$ denotes the number of belonging videos (default value is 200), $h$ represents the bandwidth of the KDE~\cite{KDEevaluation}, $G$ refers to the kernel function (default is gaussian kernel), and $t_q$ denotes the attribution signal associated with the $q$-th video. The expression $1 - \alpha$ signifies the target cumulative probability for threshold determination, where $\alpha$ is a pre-specified significance level, defaulting to 0.05. If attribution signal $t < \tau$, the test video $x$ is classified as a belonging video; otherwise, it is considered a non-belonging video.

%% file: sec/5_Exper.tex
\section{Experiments and Results}
This section first evaluates the effectiveness and efficiency of SWIFT. This is followed by an assessment of its generalization ability across models with different VAE types. Afterward, detailed ablation studies are conducted to highlight the importance of each module and parameter selection. Furthermore, we assess the robustness of SWIFT under various conditions. Additional implementation details and experiments are provided in supplementary materials.

\subsection{Experimental Setup}
\textbf{Model:} We assess the performance within 5 SOTA models equipped with 3D VAE. These models include: HunyuanVideo (HYV)~\cite{HYV}, which utilizes the ``Dual-stream to Single-stream" architecture; Wan2.1~\cite{Wan2.1}, which introduces the novel Wan-VAE and uses staged training; EasyAnimate (EA)~\cite{EA}, which integrates slice VAE with the U-ViT architecture; LTX-Video (LTX)~\cite{LTX}, a fast generative model that performs a final denoising step during decoding; and Wan2.2~\cite{Wan2.1}, the first open-source MoE-based model. These models are representative video generation models.

\noindent
\textbf{Dataset:} We constructed a dataset of 4,000 videos (named S-Video), consisting of 500 real videos randomly sampled from OpenVidHD~\cite{OpenVid} and 3,500 generated videos produced by all five models, based on 700 randomly selected prompts from OpenVidHD. For evaluation, the first 200 belonging videos from each model were used for threshold determination, while the remaining 500 videos were used for performance assessment, with the two sets being completely independent. Further details are shown in \cref{Dataset}.

\begin{table*}[]
    \centering
    \caption{Detailed composition information of the S-Video dataset.}
    \vspace{-0.2cm}
    \resizebox{0.75\linewidth}{!}{%
    \begin{tabular}{ccccccc}
    \toprule
    Dataset / Model & Specific Type & Release Date & Resolution (W$\times$H) & Frames & FPS & Quantity \\
    \midrule
    OpenVid-1M~\cite{OpenVid} & Real & 2024-07-01 & 512$\times$360$\sim$1024$\times$1024 & 30$\sim$196 & 20$\sim$60 & 500 \\
    HunyuanVideo~\cite{HYV} & T2V & 2024-12-03 & 960$\times$544 & 129 & 24 & 700 \\
    Wan2.1~\cite{Wan2.1} & T2V-1.3B & 2025-02-25 & 832$\times$480 & 81 & 16 & 700 \\
    EasyAnimate~\cite{EA} & T2V-7B-V5.1 & 2025-03-06 & 1008$\times$576 & 49 & 8 & 700 \\
    LTX-Video~\cite{LTX} & TI2V-13B-0.9.7 & 2025-05-05 & 1216$\times$704 & 121 & 30 & 700 \\
    Wan2.2~\cite{Wan2.1} & TI2V-5B & 2025-07-28 & 1280$\times$704 & 121 & 24 & 700 \\
    \bottomrule
    \end{tabular}%
    }
    \label{Dataset}
    \vspace{-0.4cm}
\end{table*}

\noindent
\textbf{Baseline:} Since the task of few-shot training-free video attribution is proposed for the first time in this paper, and no effective methods currently exist in this domain, we compared SWIFT with AEDR~\cite{AEDR}, a training-free image attribution method that currently achieves the best performance and aligns most closely with the task’s characteristics.

Our method is implemented using Python 3.10.0 and PyTorch 2.4.0. All experiments were conducted on an Ubuntu 20.04 server equipped with 8 NVIDIA RTX A6000 GPUs.

\begin{table}[]
\centering
\caption{Evaluation of SWIFT's attribution effectiveness.}
\vspace{-0.2cm}
\label{Main-1}
\resizebox{\linewidth}{!}{%
\begin{tabular}{c@{\hspace{1pt}}c@{\hspace{8pt}}c@{\hspace{3pt}}c@{\hspace{3pt}}c@{\hspace{3pt}}c@{\hspace{8pt}}cc@{\hspace{3pt}}c@{\hspace{3pt}}c@{\hspace{3pt}}c@{\hspace{8pt}}c}

\toprule
\multirow{2.5}{*}{$\mathcal{M}_1$} & \multirow{2.5}{*}{$\mathcal{M}_2$} & \multicolumn{5}{c}{AEDR~\cite{AEDR}} & \multicolumn{5}{c}{SWIFT (ours)} \\
\cmidrule(r){3-7} \cmidrule(l){8-12}
 &  & TP & FP & FN & TN & Acc & TP & FP & FN & TN & Acc \\

\midrule
\multirow{5}{*}{HYV} & Wan2.1 & 481 & 465 & 19 & 35 & 51.6\% & 473 & 6 & 27 & 494 & 96.7\%~\color[HTML]{009901}{(+45.1)} \\
 & EA & 481 & 375 & 19 & 125 & 60.6\% & 473 & 7 & 27 & 493 & 96.6\%~\color[HTML]{009901}{(+36.0)} \\
 & LTX & 481 & 467 & 19 & 33 & 51.4\% & 473 & 171 & 27 & 329 & 80.2\%~\color[HTML]{009901}{(+28.8)} \\
 & Wan2.2 & 481 & 476 & 19 & 24 & 50.5\% & 473 & 143 & 27 & 357 & 83.0\%~\color[HTML]{009901}{(+32.5)} \\
 & Real & 481 & 433 & 19 & 47 & 54.8\% & 473 & 4 & 27 & 496 & 96.9\%~\color[HTML]{009901}{(+42.1)} \\

\midrule
\multirow{5}{*}{Wan2.1} & HYV & 477 & 80 & 23 & 420 & 89.7\% & 484 & 0 & 16 & 500 & 98.4\%~\color[HTML]{009901}{(+8.7)} \\
 & EA & 477 & 45 & 23 & 455 & 93.2\% & 484 & 0 & 16 & 500 & 98.4\%~\color[HTML]{009901}{(+5.2)} \\
 & LTX & 477 & 41 & 23 & 459 & 93.6\% & 484 & 0 & 16 & 500 & 98.4\%~\color[HTML]{009901}{(+4.8)} \\
 & Wan2.2 & 477 & 155 & 23 & 345 & 82.2\% & 484 & 0 & 16 & 500 & 98.4\%~\color[HTML]{009901}{(+16.2)} \\
 & Real & 477 & 102 & 23 & 398 & 87.5\% & 484 & 0 & 16 & 500 & 98.4\%~\color[HTML]{009901}{(+10.9)} \\

\midrule
\multirow{5}{*}{EA} & HYV & 478 & 323 & 22 & 171 & 65.5\% & 478 & 0 & 22 & 500 & 97.8\%~\color[HTML]{009901}{(+32.3)} \\
 & Wan2.1 & 478 & 263 & 22 & 237 & 71.5\% & 478 & 0 & 22 & 500 & 97.8\%~\color[HTML]{009901}{(+26.3)} \\
 & LTX & 478 & 478 & 22 & 22 & 50.0\% & 478 & 0 & 22 & 500 & 97.8\%~\color[HTML]{009901}{(+47.8)} \\
 & Wan2.2 & 478 & 374 & 22 & 126 & 60.4\% & 478 & 0 & 22 & 500 & 97.8\%~\color[HTML]{009901}{(+37.4)} \\
 & Real & 478 & 296 & 22 & 204 & 68.2\% & 478 & 0 & 22 & 500 & 97.8\%~\color[HTML]{009901}{(+29.6)} \\

\midrule
\multirow{5}{*}{LTX} & HYV & 482 & 318 & 18 & 182 & 66.4\% & 474 & 195 & 26 & 305 & 77.9\%~\color[HTML]{009901}{(+11.5)} \\
 & Wan2.1 & 482 & 153 & 18 & 347 & 82.9\% & 474 & 124 & 26 & 376 & 85.0\%~\color[HTML]{009901}{(+2.1)} \\
 & EA & 482 & 11 & 18 & 489 & 97.1\% & 474 & 29 & 26 & 471 & 94.5\%~\color{red!80}{(-2.6)}\\
 & Wan2.2 & 482 & 224 & 18 & 276 & 75.9\% & 474 & 204 & 26 & 296 & 77.0\%~\color[HTML]{009901}{(+1.1)} \\
 & Real & 482 & 26 & 18 & 474 & 95.6\% & 474 & 52 & 26 & 448 & 92.2\%~\color{red!80}{(-3.4)} \\

\midrule
\multirow{5}{*}{Wan2.2} & HYV & 474 & 179 & 26 & 321 & 79.5\% & 483 & 14 & 17 & 486 & 96.9\%~\color[HTML]{009901}{(+17.4)} \\
 & Wan2.1 & 474 & 371 & 26 & 129 & 60.3\% & 483 & 0 & 17 & 500 & 98.3\%~\color[HTML]{009901}{(+38.0)} \\
 & EA & 474 & 205 & 26 & 295 & 76.9\% & 483 & 0 & 17 & 500 & 98.3\%~\color[HTML]{009901}{(+21.4)} \\
 & LTX & 474 & 147 & 26 & 353 & 82.7\% & 483 & 3 & 17 & 497 & 98.0\%~\color[HTML]{009901}{(+15.3)} \\
 & Real & 474 & 44 & 26 & 456 & 93.0\% & 483 & 3 & 17 & 497 & 98.0\%~\color[HTML]{009901}{(+5.0)} \\

\midrule
\rowcolor{gray!25} \textbf{Avg Acc} & \multicolumn{1}{l}{} & \multicolumn{1}{l}{} & \multicolumn{1}{l}{} & \multicolumn{1}{l}{} & \multicolumn{1}{l}{} & \textbf{73.6\%} & \multicolumn{1}{l}{} & \multicolumn{1}{l}{} & \multicolumn{1}{l}{} & \multicolumn{1}{l}{} & \textbf{94.0\%~\color[HTML]{009901}{(+20.4)}} \\
\bottomrule

\end{tabular}%
}
\vspace{-0.6cm}
\end{table}

\subsection{Effectiveness}
We first evaluated the attribution effectiveness of SWIFT on the S-Video dataset, which measures whether a given test video was generated by the target model (binary classification). The experimental results are summarized in \cref{Main-1}, where $\mathcal{M}_1$ denotes the target model and $\mathcal{M}_2$ refers to other model. Across all five models in S-Video, SWIFT achieved an average attribution accuracy of 94.0\%, substantially outperforming AEDR’s~\cite{AEDR} 73.6\%. Furthermore, SWIFT maintained exceptionally high attribution accuracy with remarkably low false positive rates (FPRs) on Wan2.1, Wan2.2~\cite{Wan2.1}, and EA~\cite{EA}, while AEDR exhibited considerably higher FPRs on these models—indicating insufficient separation between belonging and non-belonging videos. Although SWIFT demonstrated outstanding and consistent attribution performance on these three models, its accuracy fluctuated to varying degrees on HYV~\cite{HYV} and LTX~\cite{LTX}.

For HYV, the performance declined when discriminating videos generated by LTX and Wan2.2, compared with other models and real videos. This drop can be attributed to HYV’s training procedure, which employed a large $L1$ loss term in its VAE, emphasizing fine-grained detail reconstruction. In contrast, both LTX and Wan2.2 used higher spatial–temporal compression ratios (see \cref{Generalization}), resulting in generated videos with reduced fidelity in spatial details. This blurring of fine details hindered SWIFT’s ability to effectively capture temporal characteristics during reconstruction. This is particularly problematic since temporal dependencies are expressed through the reconstruction loss, which in turn impacts attribution accuracy. Moreover, because LTX applied an aggressive sampling strategy, its initial generation quality was relatively poor. To compensate, LTX not only performed up sampling during decoding but also introduced a denoising step. While denoising enhanced the perceived visual quality, it also substantially reduced the intrinsic discrepancy between the original and reconstructed videos, diminishing SWIFT’s sensitivity to temporal cues. Nevertheless, LTX still achieved over 90\% attribution accuracy in distinguishing belonging videos from EA and Real videos, indicating that such fluctuations are more pronounced among structurally similar models.

Despite occasional performance variations in specific cases, SWIFT consistently outperformed AEDR across all five models. This superiority stems from SWIFT’s deep mining of the video's temporal characteristics, enabling it to more comprehensively capture the intrinsic distinctions between belonging and non-belonging videos, thus delivering more accurate and robust attribution results.


\begin{table}[]
\centering
\caption{Running time on different models.}
\vspace{-0.2cm}
\resizebox{\columnwidth}{!}{%
\begin{tabular}{c@{\hspace{8pt}}c@{\hspace{6pt}}c@{\hspace{6pt}}c@{\hspace{6pt}}c@{\hspace{6pt}}c}
\toprule
Model & HYV~\cite{HYV} & Wan2.1~\cite{Wan2.1} & EA~\cite{EA} & LTX~\cite{LTX} & Wan2.2~\cite{Wan2.1} \\
\midrule
Precision & FP16 & FP32 & FP16 & BF16 & FP32 \\
Generation & 2369s & 349s & 421s & 39s & 656s \\
AEDR & 227s & 46s & 31s & 17s & 186s \\
SWIFT & 218s~\color[HTML]{009901}{(-9s)} & 33s~\color[HTML]{009901}{(-13s)} & 21s~\color[HTML]{009901}{(-10s)} & 14s~\color[HTML]{009901}{(-3s)} & 170s~\color[HTML]{009901}{(-16s)} \\
\rowcolor{gray!25} 
\textbf{Speedup} & \textbf{4\%} & \textbf{28\%} & \textbf{32\%} & \textbf{18\%} & \textbf{9\%} \\
\bottomrule
\end{tabular}%
}
\label{Time}
\vspace{-0.5cm}
\end{table}

\subsection{Efficiency}
We evaluate the computational efficiency of SWIFT. Comprehensive tests were conducted on all models, each evaluated on 500 belonging videos, and the average runtime was measured for comparison. All experiments were performed using the recommended precision. The experimental results are presented in \cref{Time}. Due to the fact that AEDR requires two reconstructions of the entire video, whereas SWIFT performs reconstruction using video windows, it achieved a performance improvement ranging from 4\% to 32\% compared to AEDR. Across the five models, SWIFT’s runtime ranged from 14s to 218s, with attribution time decreasing notably as the window size was reduced (see \cref{Length}).

\begin{table*}[t]
\centering
\caption{The generalization capability of SWIFT across different types and functions of 3D VAE.}
\vspace{-0.2cm}
\resizebox{0.8\textwidth}{!}{%
\begin{tabular}{c|cccc|c}
\toprule
Model & HYV~\cite{HYV} & Wan2.1~\cite{Wan2.1} & EA~\cite{EA} & Wan2.2~\cite{Wan2.1} & LTX~\cite{LTX} \\
\midrule
Spatial–Temporal Compression & 8$\times$8$\times$4 & 8$\times$8$\times$4 & 8$\times$8$\times$4 & 16$\times$16$\times$4 & 32$\times$32$\times$8 \\
Number of Output Channels & 16 & 16 & 16 & 48 & 128 \\
Total Compression & 48 & 48 & 48 & 64 & 192 \\
Functions of VAE & \multicolumn{4}{c|}{Up and Down Sampling} & \makecell{Up and Down Sampling \\ \& Final Denoising Step} \\
\rowcolor{gray!25}
\textbf{SWIFT} & \textbf{90.7\%~\color[HTML]{009901}{}} & \textbf{98.4\%~\color[HTML]{009901}{}} & \textbf{97.8\%~\color[HTML]{009901}{}} & \textbf{97.9\%~\color[HTML]{009901}{}} & \textbf{85.3\%~\color[HTML]{009901}{}} \\
\bottomrule
\end{tabular}%
}
\label{Generalization}
\vspace{-0.5cm}
\end{table*}

\subsection{Generalization}
This section evaluates the generalization ability of SWIFT across video generation models equipped with different types of 3D VAEs, with the results summarized in \cref{Generalization}. Specifically, the VAEs of HYV~\cite{HYV}, Wan2.1, and EA~\cite{EA} adopt a spatial–temporal compression ratio of 8$\times$8$\times$4, whereas Wan2.2~\cite{Wan2.1} employs a stronger compression ratio (16$\times$16$\times$4), indicating a higher level of compression. The VAEs of these four models solely function as mappings between the pixel space and the latent space. In contrast, the VAE of LTX~\cite{LTX} incorporates an additional denoising step during decoding to mitigate the degradation in video quality caused by its excessively high compression ratio (32$\times$32$\times$8). This design partially improves video visual quality and narrows the gap between the original and reconstructed videos. However, such a compensatory effect simultaneously reduces SWIFT’s sensitivity to temporal discrepancies, leading to a slight drop in attribution accuracy to 85.3\% on LTX~\cite{LTX}. Despite this, SWIFT achieves an average attribution accuracy of 94.0\% across all five models, demonstrating strong generalization ability and robust performance across diverse types of video generation models.

\begin{table}[]
\centering
\caption{Attribution accuracy of different loss metrics.}
\vspace{-0.2cm}
\resizebox{0.7\columnwidth}{!}{%
\begin{tabular}{c|cccc|c}
\toprule
Loss Metric & TP & FP & FN & TN & Acc \\
\midrule
MAE & 483 & 495 & 17 & 5 & 97.8\% \\
\rowcolor{gray!25} 
MSE & 484 & 500 & 16 & 0 & \textbf{98.4}\% \\
PSNR & 478 & 0 & 22 & 500 & 47.8\% \\
SSIM~\cite{SSIM} & 470 & 1 & 30 & 499 & 47.1\% \\
LPIPS~\cite{LPIPS} & 479 & 499 & 21 & 1 & 97.8\% \\
\bottomrule
\end{tabular}%
}
\label{Loss Metric}
\vspace{-0.5cm}
\end{table}

\vspace{-0.15cm}
\subsection{Ablation Studies}
\vspace{-0.05cm}
\noindent
\textbf{Loss Metric:} We used Wan2.1~\cite{Wan2.1} as the target model to investigate the impact of different loss metrics on attribution accuracy. In the experiment, 500 belonging videos and 500 non-belonging videos (randomly selected from both generated and real videos) were tested. We compared five commonly used metrics: Mean Absolute Error (MAE), Mean Squared Error (MSE), Peak Signal-to-Noise Ratio (PSNR), Structural Similarity Index (SSIM)~\cite{SSIM}, and Learned Perceptual Image Patch Similarity (LPIPS)~\cite{LPIPS}. The results, presented in \cref{Loss Metric}, reveal that the highest accuracy of 98.4\% was achieved using MSE. While the existing models allocate a relatively high proportion to the $L1$ loss term during VAE training, leading to an accuracy of 97.8\% for MAE, MSE outperforms MAE as it more effectively amplifies differences, thus providing the best performance.


\noindent
\textbf{Window Selection:} The selection of windows is determined by both the temporal compression ratio $K$ and the generative characteristics of the target model. Consequently, we select two representative models: HYV~\cite{HYV} and LTX~\cite{LTX}. For HYV ($K=4$), the optimal performance is achieved by selecting $W_0$ and $W_3$, which is validated in \cref{Window} and aligns with the theoretical analysis in \cref{3-3}. For LTX ($K=8$), while $W_0$ and $W_7$ appear to be the more intuitive choice, the combination of $W_0$ and $W_3$ proves superior due to the VAE's unique decoder incorporating a denoising step. Crucially, this does not deviate from the theoretical framework of “maximizing the reconstruction discrepancy.” By quantifying the divergence between various corrupted and normal windows, we demonstrate that the $\{W_0, W_3\}$ pair actually induces the most significant disruption (see supplementary material). Specifically, for models employing VAEs with purely compression-based functionality, the reconstructed videos retain pronounced VAE distributional signatures. In such cases, the $\{W_0, W_{K-1}\}$ pair serves as a reliable approximation for the optimal solution. For models utilizing VAEs that integrate both compression and denoising, the denoising process during decoding substantially attenuates these inherent distributional traits. This necessitates precise quantification to identify the specific window combination that yields the maximum discrepancy.

\begin{table}[]
\centering
\caption{Attribution accuracy of different window selection.}
\vspace{-0.2cm}
\resizebox{\columnwidth}{!}{%
\begin{tabular}{c||c|c|c||c|c|c}
\toprule
Model & Normal & Corrupted & Acc & Normal & Corrupted & Acc \\
\midrule
\multirow{3}{*}{\makecell{HYV \\ $K=4$}} & \multirow{3}{*}{$W_0$} & $W_1$ & 82.3\% & \multirow{3}{*}{$W_4$} & $W_1$ & 73.3\% \\
 &  & $W_2$ & 82.3\% &  & $W_2$ & 72.4\% \\
 &  & $W_3$ & \cellcolor{gray!25}\textbf{90.7\%} &  & $W_3$ & \cellcolor{gray!25}88.0\% \\
\midrule
\multirow{7}{*}{\makecell{LTX \\ $K=8$}} & \multirow{7}{*}{$W_0$} & $W_1$ & 81.6\% & \multirow{7}{*}{$W_8$} & $W_1$ & 49.5\% \\
 &  & $W_2$ & 76.8\% &  & $W_2$ & 50.1\% \\
 &  & $W_3$ & \cellcolor{gray!25}\textbf{85.3\%} &  & $W_3$ & 53.4\% \\
 &  & $W_4$ & 73.2\% &  & $W_4$ & 52.8\% \\
 &  & $W_5$ & 81.6\% &  & $W_5$ & \cellcolor{gray!25}54.4\% \\
 &  & $W_6$ & 75.6\% &  & $W_6$ & 51.7\% \\
 &  & $W_7$ & 78.6\% &  & $W_7$ & 53.9\% \\
\bottomrule
\end{tabular}%
}
\label{Window}
\vspace{-0.3cm}
\end{table}

\begin{table}[]
\centering
\caption{Impact of sample size on attribution accuracy.}
\vspace{-0.2cm}
\resizebox{1\columnwidth}{!}{%
\begin{tabular}{ccccccc}
\toprule
\multirow{2.5}{*}{$S$} & \multicolumn{5}{c}{Attribution Accuracy} & \multirow{2}{*}{Avg Acc} \\
\cmidrule(r){2-6}
 & HYV & Wan2.1 & EA & LTX & Wan2.2 &  \\
\midrule
\rowcolor{gray!25}
0 & \textbf{90.1\%} & 81.7\% & \textbf{89.3\%} & 66.1\% & \textbf{98.4\%} & 85.1\% \\
5 & 78.6\% & 88.7\% & 80.3\% & 79.0\% & 91.8\% & 83.7\% \\
10 & 76.1\% & 88.3\% & 80.8\% & 78.3\% & 90.4\% & 82.8\% \\
\rowcolor{gray!25}
\textbf{20} & 85.2\% & 94.2\% & 95.0\% & 83.9\% & 92.7\% & \textbf{90.2\%} \\
35 & 90.3\% & 96.9\% & 97.4\% & 84.8\% & 93.2\% & 92.5\% \\
50 & 90.7\% & 97.1\% & 96.6\% & 84.7\% & 93.3\% & 92.5\% \\
100 & 90.2\% & 98.4\% & 97.7\% & 85.0\% & 97.4\% & 93.7\% \\
150 & 90.2\% & 98.4\% & 97.8\% & 85.1\% & 97.6\% & 93.8\% \\
\rowcolor{gray!25}
200 & 90.7\% & 98.4\% & 97.8\% & 85.3\% & 97.9\% & \textbf{94.0\%} \\
\bottomrule
\end{tabular}%
}
\label{Samples}
\vspace{-0.5cm}
\end{table}

\noindent
\textbf{Determining Threshold Sample Size:} We investigated the impact of the sample size ($S$) used to determine the threshold on attribution accuracy, conducting tests across all five models. The results are presented in \cref{Samples}. It was observed that as the sample size $S$ increased from 5 to 200, the attribution accuracy gradually improved, reaching its peak at 94\% when $S=200$. Notably, the average accuracy already reached 90\% at $S=20$, which demonstrates that SWIFT possesses few-shot attribution capabilities. From the analysis in \cref{Introduction}, the attribution signal for the belonging videos tends to be significantly less than 1, whereas that for non-belonging videos tends to be close to 1. By setting the threshold $\tau=1$, the HYV, EA~\cite{EA}, and Wan2.2~\cite{Wan2.1} achieved accuracies of 90.1\%, 89.3\%, and 98.4\%, validating SWIFT's zero-shot attribution capability for certain models. These experimental results confirm that SWIFT is highly valuable, as it enables reliable attribution even with a restricted-access scenario, requiring only 20 samples.

\noindent
\textbf{Selection of KDE Parameters.} To evaluate the impact of bandwidth and kernel function on attribution accuracy, we used the EA~\cite{EA} as the target model and tested it across the S-video. We examined two common bandwidth selection methods, Scott and Silverman, along with three widely used kernel functions: Gaussian, Uniform, and Epanechnikov. \Cref{KDE} shows that the choice of bandwidth and kernel function has slight impact on accuracy, with fluctuations under 0.5\%. Therefore, we adopted Scott for bandwidth and the Gaussian kernel as the default configuration, as this combination yielded the best overall performance.

\begin{table}[]
\centering
\caption{Impact of bandwidth and kernel function on accuracy.}
\vspace{-0.2cm}
\resizebox{0.63\columnwidth}{!}{%
\begin{tabular}{c|c|c}
\toprule
Bandwidth & Kernel Function & Avg Acc \\
\midrule
\multirow{3}{*}{Scott} & Gaussian & \cellcolor{gray!25}\textbf{97.80\%} \\
 & Uniform & 97.40\% \\
 & Epanechnikov & 97.40\% \\
\midrule
\multirow{3}{*}{Silverman} & Gaussian & \cellcolor{gray!25}\textbf{97.80\%} \\
 & Uniform & 97.30\% \\
 & Epanechnikov & 97.40\% \\
\bottomrule
\end{tabular}%
}
\label{KDE}
\vspace{-0.3cm}
\end{table}

\begin{table}[]
\centering
\caption{Impact of video length on accuracy and runtime.}
\vspace{-0.2cm}
\resizebox{0.63\columnwidth}{!}{%
\begin{tabular}{cccc}
\toprule
Length & Frames &  Runtime & Accuracy \\
\midrule
\rowcolor{gray!25}
\textbf{100\%} & 121 & 170s & \textbf{98.0\%} \\
75\% & 93 & 134s & 98.0\% \\
50\% & 61 & 84s & 97.1\% \\
\rowcolor{gray!25}
\textbf{25\%} & 33 & \textbf{40s} & 94.1\% \\
\bottomrule
\end{tabular}%
}
\label{Length}
\vspace{-0.5cm}
\end{table}

\noindent
\textbf{Reconstruction Window Size:} To maximize the visual information captured within the windows, we defaulted to applying the SWIFT to the entire video. To evaluate both time performance and robustness, we selected Wan2.2~\cite{Wan2.1} as the target model and tested the effect of belonging video length on attribution accuracy and runtime. The results are shown in \cref{Length}. As the video length decreased from 100\% to 25\%, the attribution accuracy slightly decreased from 98.0\% to 94.1\%, while the algorithm runtime significantly dropped from 170s to 40s. These results show the strong robustness of SWIFT to variations in video length. In resource-constrained situations, shorter videos can be used for fast attribution without significantly affecting performance.

\vspace{-0.15cm}
\subsection{Robustness}
\vspace{-0.05cm}
\label{Robustness}
To test the robustness of SWIFT against common post-processing operations, we applied central cropping to 50\% (Crop), H.264 compression with CRF=28 (Compression), random horizontal and vertical flips (Flip), and Gaussian noise ($\mu=0$, $\sigma=0.05$, Noise). We used EA~\cite{EA} as the target model and tested it across the S-Video dataset. The results, shown in \cref{Rob}, indicate that SWIFT’s attribution accuracy decreased by 9.5\% to 45.8\%. Among the applied operations, Flip and Noise had the most significant impact on attribution, as they substantially altered the feature distribution and visual quality of the test video, introducing reconstruction bias. Nevertheless, despite the performance degradation induced by these four operations, SWIFT consistently maintained a superior attribution accuracy compared to AEDR~\cite{AEDR} throughout the entire evaluation.

\begin{table}[]
\centering
\caption{Performance of SWIFT on post-processing operations.}
\vspace{-0.2cm}
\resizebox{\columnwidth}{!}{%
\begin{tabular}{c@{\hspace{6pt}}c@{\hspace{6pt}}c@{\hspace{6pt}}c@{\hspace{6pt}}c@{\hspace{6pt}}c}
\toprule
Method & w/o & Crop & Compression & Flip & Noise \\
\midrule
AEDR & 63.1\% & 37.2\% & 42.6\% & 29.4\% & 49.9\% \\
\rowcolor{gray!25}
SWIFT & 97.8\% & 81.9\%~\color[HTML]{009901}{(+44.7)} & 88.3\%~\color[HTML]{009901}{(+45.7)} & 52.0\%~\color[HTML]{009901}{(+22.6)} & 60.3\%~\color[HTML]{009901}{(+10.4)} \\
\bottomrule
\end{tabular}%
}
\label{Rob}
\vspace{-0.5cm}
\end{table}

%% file: sec/6_Limitation.tex
\vspace{-0.2cm}
\section{Limitations and Future Work}
\vspace{-0.1cm}
Although SWIFT achieves high attribution accuracy for most video generation models that utilize 3D VAEs, its performance slightly diminishes when applied to VAE architectures incorporating denoising mechanisms. However, such models are relatively rare. Moreover, SWIFT is not directly applicable to early video and image generation models, as these models do not manipulate the temporal dimension in their 2D VAEs, thus preventing the establishment of temporal mappings. As video generation progresses toward higher resolutions and longer durations, the use of 3D VAEs becomes increasingly essential for reducing computational load, making SWIFT more relevant for future applications. Currently, the robustness of training-free attribution methods remains suboptimal, and future research could focus on enhancing the robustness of these methods.



%% file: sec/7_Conclusion.tex
\vspace{-0.2cm}
\section{Conclusion}
\vspace{-0.1cm}

In this paper, we define the ``few-shot training-free generated video attribution" task for the first time and propose a solution tightly integrated with the temporal characteristics of videos (namely SWIFT). SWIFT exploits the temporal mapping of ``Pixel Frames(many)$\leftrightarrow$Latent Frame(one)" within each video chunk. It applies a sliding window to perform two distinct reconstructions (normal and corrupted). The attribution signal is derived from the average loss ratio of overlapping frames between the two reconstructions, achieving high attribution accuracy. Experimental evaluations demonstrate that SWIFT attains over 90\% attribution accuracy with merely 20 video samples across all five models and even enables zero-shot attribution for models such as HunyuanVideo, EasyAnimate, and Wan2.2.

%% file: sec/8_Ack.tex
\section{Acknowledgments}
This work was supported in part by the National Natural Science Foundation of China (Grants 62472398, U2336206, and 62302146), the New Generation Artificial Intelligence-National Science and Technology Major Project (No. 2025ZD0123202), and the Fundamental Research Funds for the Central Universities of China (Grant PA2025IISL0104).